\documentclass[letterpaper, 10 pt, conference]{ieeeconf}  % Comment this line out if you need a4paper

\IEEEoverridecommandlockouts                              % This command is only needed if 
\title{\LARGE \bf
HPRM: High-Performance Robotic Middleware for Intelligent Autonomous Systems}

\author{Jacky Kwok, Shulu Li, Marten Lohstroh, Edward A. Lee% <-this % stops a space
}

\usepackage[hidelinks]{hyperref}
\usepackage{graphicx}
\usepackage{amsmath, amssymb}
\usepackage{listings}
\usepackage{xcolor} % Required for custom colors
\usepackage{url}

\definecolor{keywordcolor1}{HTML}{A048DC}    % Purple
\definecolor{keywordcolor2}{HTML}{55B8BF}   % Light Blue
\definecolor{keywordcolor3}{HTML}{E2931D}        % Orange
\definecolor{commentcolor}{HTML}{90A4AE}         % Green
\definecolor{stringcolor}{rgb}{0.9,0,0}          % Red
\definecolor{backcolour}{rgb}{1,1,1}             % White
\definecolor{codegray}{rgb}{0.5,0.5,0.5}         % Gray for line numbers

\lstdefinestyle{linguafrancastyle}{
    backgroundcolor=\color{backcolour},
    commentstyle=\color{commentcolor},
    basicstyle=\footnotesize\ttfamily,
    breakatwhitespace=false,         
    breaklines=true,
    captionpos=b,
    keepspaces=true,                 
    numbers=left,                    
    numbersep=5pt,
    showspaces=false,                
    showstringspaces=false,
    showtabs=false,                  
    tabsize=2,
    framexleftmargin=15pt,
    xleftmargin=15pt,
    frame=tlbr,
    framesep=5pt,
    framerule=0pt,
    language=[LaTeX]TeX,
    morekeywords=[1]{reactor, input, output, int, serializer},
    morekeywords=[2]{reaction, target, new, deadline, after},
    morekeywords=[3]{A, B, startup},
    literate={->}{{{\color{keywordcolor2}->}}}2,
    keywordstyle=[1]\color{keywordcolor1},
    keywordstyle=[2]\color{keywordcolor2},
    keywordstyle=[3]\color{keywordcolor3},
    morecomment=[l]{\#},
    morestring=[b]",
}

\begin{document}

\maketitle
\thispagestyle{empty}
\pagestyle{empty}

%%%%%%%%%%%%%%%%%%%%%%%%%%%%%%%%%%%%%%%%%%%%%%%%%%%%%%%%%%%%%%%%%%%%%%%%%%%%%%%%
\begin{abstract}
The rise of intelligent autonomous systems, especially in robotics and autonomous agents, has created a critical need for robust communication middleware that can ensure real-time processing of extensive sensor data. Current robotics middleware like Robot Operating System (ROS) 2 faces challenges with nondeterminism and high communication latency when dealing with large data across multiple subscribers on a multi-core compute platform. To address these issues, we present High-Performance Robotic Middleware (HPRM), built on top of the deterministic coordination language Lingua Franca (LF). HPRM employs optimizations including an in-memory object store for efficient zero-copy transfer of large payloads, adaptive serialization to minimize serialization overhead, and an eager protocol with real-time sockets to reduce handshake latency. Benchmarks show HPRM achieves up to 173x lower latency than ROS2 when broadcasting large messages to multiple nodes. We then demonstrate the benefits of HPRM by integrating it with the CARLA simulator and running reinforcement learning agents along with object detection workloads. In the CARLA autonomous driving application, HPRM attains 91.1\% lower latency than ROS2. The deterministic coordination semantics of HPRM, combined with its optimized IPC mechanisms, enable efficient and predictable real-time communication for intelligent autonomous systems. 
% Videos and code to reproduce our results are available at \href{https://depetrol.github.io/HPRM}{https://depetrol.github.io/HPRM}.
\end{abstract}

%%%%%%%%%%%%%%%%%%%%%%%%%%%%%%%%%%%%%%%%%%%%%%%%%%%%%%%%%%%%%%%%%%%%%%%%%%%%%%%%
\section{Introduction}

Due to the advancements in AI, the area of intelligent autonomous systems is rapidly growing. These systems, especially in the context of robotics and autonomous agents, are critical in both performance and reliability due to their capability to analyze extensive sensor data in real-time. They require a robust communication infrastructure to ensure real time transmission and processing of data.

In the architecture of autonomous systems, modules are typically organized as coarse-grained processes~\cite{sifakis2019autonomous}. This design paradigm, which emphasizes functional independence and resource isolation, ensures that a failure in one module does not compromise the integrity or functionality of other modules or the system as a whole. Consequently, the exchange of data across different modules is predominantly facilitated through Inter-Process Communication (IPC) \cite{dinari2020inter, venkataraman2015evaluation} techniques. For instance, in a scenario where a robot is tasked with identifying specific objects for humans, the image data captured by the camera module undergoes several steps: serialization into a buffer, copying into the system kernel, transferring to the target process, and finally, deserialization. These operations usually lead to high latency in applications that make use of high-resolution cameras or LiDAR sensors.

Frameworks like the Robot Operating System (ROS) \cite{quigley2009ros} and MQTT \cite{soni2017survey} have seen significant adoption in critical, concurrent, and distributed settings, including autonomous vehicles and industrial automation. These frameworks are valued for their convenience, modularity, and the use of a publish-subscribe mechanism, which can easily be leveraged for message exchange in distributed systems. However, the publish-subscribe mechanism, particularly in high-stakes environments like autonomous driving, introduces a level of nondeterminism \cite{hunt2013ddos} due to varying communication timing, potentially resulting in unpredictable message handling sequences. This unpredictability is a significant concern in environments where the consequences of errors are severe. In addition, to support message passing in autonomous navigation systems, frameworks such as ROS2 generally utilize sockets-based communication \cite{blass2021automatic}. However, this method falls short in scenarios involving the processing of large data packets across numerous subscribers, as it leads to an increase in communication latency with message size.

In this study, we offer an alternative to ROS2—High-Performance Robotic Middleware (HPRM). HPRM is an open-source robotic middleware built on top of a coordination language, Lingua Franca (LF) \cite{lohstroh2021toward}. LF, which is based on the reactor model \cite{lohstroh2020reactors}, is a polyglot coordination language that combines the most effective semantic elements from well-established computational models. This includes the actor model, Logical Execution Time (LET), synchronous reactive languages, and discrete event systems like SystemC. LF advances the field by integrating time as a primary element within its programming paradigm, thereby facilitating deterministic interactions across various physical and logical timelines. HPRM is meant to enhance the capability of robotic middleware in handling large volumes of sensor data and ML workloads using efficient IPC techniques. Specifically, it uses an in-memory object store to efficiently transfer large objects across different processes, adaptive serialization for different types of sensor data in Python, and an eager protocol and real-time sockets to minimize the handshake latency for transmitting control and object references. Our approach significantly reduces the overhead associated with local IPC compared to ROS2 \cite{puck2021performance}.

We demonstrate the benefits of HPRM by integrating it with CARLA \cite{dosovitskiy2017carla} to be running reinforcement learning (RL) agents and object detection in autonomous driving scenarios. ROS applications can also be easily ported to HPRM, and a demonstration of porting ROS2 programs to HPRM has been included in the GitHub repository and showcased in the video. The HPRM runtime system is implemented in C and applications are modular, just like ROS and MQTT, allowing independent processes to be deployed.

\section{Motivation and Requirements}

\subsection{Motivation}

Frameworks like ROS2, are becoming more prevalent in critical applications, including autonomous driving, where the implications of unpredictable behaviors are significant. However, the coordination mechanism in ROS2 introduces nondeterminism~\cite{hunt2013ddos}, leading to arbitrary ordering in the handling of messages. This inherent nondeterminism in the publish-subscribe communication models poses a risk and could compromise the reliability of such systems.

Furthermore, robotics middleware \cite{fitzpatrick2014middle}, such as ROS2, faces considerable delays in message delivery, which can compromise the efficacy of real-time robotic operations when dealing with large volumes of data or multiple subscribers. Consider the scenario in autonomous navigation systems, where a planning module has to process large-scale inputs from perception  before sending actions to other components like a localization module and vehicle control systems. Moreover, many developers within the ROS community have experienced latency problems when publishing large data \cite{ros2_rosidl_issue_156}\cite{ros2_rclpy_issue_763}\cite{ros2_issue_1242}. Therefore, minimizing communication delays is pivotal for improving the real-time responsiveness of robotic systems, thereby improving the overall user experience in scenarios that demand real-time data processing. 

Kronaur et al. \cite{kronauer2021latency} highlights the proportional increase in communication latency relative to message size using ROS2. Specifically, they observe that for messages around 4MB, the median delay for ROS2 is around 10ms. In scenarios of 1MB data being distributed to five subscribers, ROS2 exhibited a median latency nearing 80ms.

% \subsection{Requirements}
% This section outlines the key features that HPRM must possess to address the challenges faced by current robotics middleware. The three main requirements for HPRM are:

% \begin{itemize}
% \item \textbf{Efficient Communication:} HPRM must prioritize minimal data movement and employ zero-copy mechanisms wherever possible to maintain low communication latency, even as message sizes increase. 
% \item \textbf{Real-time Performance and Fault Tolerance:} HPRM must guarantee real-time properties, maintaining high reliability and successful message delivery even in scenarios with heavy workloads and multiple subscribers. The framework should include mechanisms for detecting and handling faults caused by violations of timing requirements, allowing for application-specific fault handlers.
% \item \textbf{Ease of Use and Integration:} HPRM should abstract away the complexities of managing object read and write access, as well as the decision-making process for selecting appropriate transport mechanisms based on message size and type. The framework should be designed for integration with popular ML frameworks (TensorFlow, PyTorch<Citations ToDO>) and ensures easy portability from existing robotics middleware like ROS2, requiring minimal modifications. Additionally, HPRM should enable users to easily deploy their applications across various embedded platforms without concern for platform-specific details.
% \end{itemize}

\section{ROS2 vs. Lingua Franca}

\subsection{ROS2}

ROS2 is designed to support robotics application development. It allows developers to encapsulate software components within distinct units known as nodes, each running within its own OS process. Nodes engage through a publisher-subscriber (pub-sub) system, where publishers announce topics and subscribers associate specific callback functions with those topics. In this paper, we employ ROS2, which utilizes a communication framework compliant with the Data Distribution Service (DDS) \cite{pardo2003omg} to facilitate the pub-sub mechanism. 

The zero-copy feature has been incorporated into both Cyclone DDS and Fast DDS. However, this feature is currently only available for rclcpp, the C++ implementation of the ROS2 client library. As of now, there is no support for the ROS2 Python client library ~\cite{ros2_rclpy_issue_833}.

When ROS2 nodes are running on the same hardware, a NIC-level loopback is applied without any network transmission \cite{liu2020robotic, liu2022zoro}. In such methods, messages are copied several times throughout processes and OS-kernel levels, leading to unnecessary memory copy and system calls. Furthermore, since a socket is a point-to-point communication interface, collective communications become inefficient. For example, if one process publishes the same message to the other three processes, the entire communication stream is repeated three times.

Wang, et al. propose a hybrid solution termed Towards Zero Copy (TZC)\cite{wang2019tzc}, designed to optimize the handling of large messages in ROS2. In TZC, messages are separated; a lightweight descriptor traverses the conventional path over a ROS topic via TCPROS, while the main body of the message resides in shared memory. However, TZC's lack of a robust mechanism to manage message lifecycles could potentially leave unclaimed payloads, risking memory leaks if their descriptors fail to be accurately transmitted. Furthermore, TZC is not actively developed or maintained and is not compatible with ROS2.

\subsection{Lingua Franca}

Lingua Franca (LF) is presented as an open-source polyglot coordination language designed to facilitate deterministic interactions among concurrent and reactive components known as reactors. The characteristic of LF that underpins our research is its deterministic nature~\cite{lee2023consistency}. LF orchestrates event flow through a system, where events are tagged, facilitating transmission from one reactor's port to another's. Each event is marked with a logical tag from a totally-ordered set G, ensuring every reactor processes events in a sequential tag order. Each event tag consists of a timestamp $t \in T$ indicating logical time and a microstep $m \in N$ for capturing super-dense time, allowing for precise event scheduling.

LF's design supports polyglot programming, enabling reactions within reactors to be authored in a variety of programming languages, including C, C++, Python, TypeScript, or Rust. This polyglot capability ensures that LF can be seamlessly integrated into diverse development environments by compiling LF programs into the chosen target language \cite{menard2023high}. HPRM is developed on top of the Python target and will support C++ in the future. Currently, LF's Python and C-runtime support various embedded platforms, including Arduino, Raspberry Pi, and Zephyr RTOS.

Furthermore, for extremely latency-critical tasks, the optimal
solution would be to avoid serialization through intra-process
communication, allowing direct access to messages without
copying or serialization. HPRM allows users to
easily switch to intra-process communication in Python. Kwok et al. ~\cite{kwok2024efficient} have enabled users to write truly
concurrent Python programs without the limitations imposed
by the Global Interpreter Lock. The syntax and documentation of LF is available at \href{https://www.lf-lang.org}{https://www.lf-lang.org}.

Time is treated as the core element in LF, with the framework providing access to both logical and physical clocks. The design principle is such that logical time closely follows physical time, maintaining a temporal coherence that ensures logical events occur near their physical counterparts but not before. Reactions can be assigned deadlines, and LF supports deadline handlers for managing situations where deadlines are violated, thereby maintaining system responsiveness and reliability. Reactor A has a specified deadline of 10 milliseconds (this value can be adjusted as a parameter of the reactor). If the reaction to event x is triggered for more than 10 milliseconds in physical time, the fault handler code will be executed in place of the first body of code.

LF employs socket-based IPC methods. However, such a socket communication mechanism is not satisfactory for the processing of large-scale sensor data or machine learning (ML) workloads, and communication latency would increase with the growth of the message size.

Our contribution through this paper is the expansion of LF's deterministic properties, enabling efficient IPC between reactors while preserving determinism. 

\section{Coordination and Optimizations}

\subsection{Centralized Coordination}

In HPRM, we use a centralized coordination mechanism. The Runtime Infrastructure (RTI) is employed to manage communication and synchronization among distributed components, named federates. In this strategy, the RTI is responsible for monitoring and regulating event tags during advancement of logical time, thereby assuring that federates process messages in a global consistent order. The RTI keeps track of the information below for each federate, identified as $f$:
\begin{itemize}
    \item \textbf{Tag Advance Grant ($TAG_f$)}: The latest tag sent to federate  $f$, enabling it to update its current event tag to $TAGf$. Initially,  $TAGf$ is set to $- \infty$.
    \item \textbf{Logical Tag Complete ($LTC_f$)}: This represents the most recent tag reported by federate $f$, signifying the completion of all tasks (computations and communications) associated with that tag or any preceding it.
    \item \textbf{Next Event Tag ($NET_f$)}: This indicates the latest event tag from federate  $f$, essentially the earliest future event in its queue. An empty queue is denoted by a special maximal tag, $\infty$. Absence of an $NET$ message would be represented as $- \infty$.
\end{itemize}

For a federate $p$ to advance to a logical time $t$ in response to its upstream reaction, it must first receive authorization from the RTI. This authorization is contingent upon the RTI's assurance that $p$ has received all messages up to and including time $t$.

A fundamental rule in this model is that a federate's logical time does not precede the physical time as indicated by its local physical clock. 
\[
\text{s.out} \rightarrow \text{p.in\:after\:200\:msec};
\]

In the connection above, a message with timestamp $t$ from sender $s$ cannot be sent before the local clock at $s$ reaches $t$ and also cannot be sent before the RTI grants to $s$ a time advance to $t$. It is noted that given that $s$ lacks upstream federates, the RTI always grants it a time advance.

If we denote the communication latency as $L$, the message from $s$ to $p$ will reach $p$ only after physical time $t+L$ measured by $s$'s physical clock. If there is a clock discrepancy $E$ between $s$'s and $p$'s hosts, $p$ will receive the message at physical time $t+E+L$ measured by $t$'s physical clock. The delay parameter $a$ ($200$ msec in the example) in the after clause then determines the timestamp $t+a$ for the message as received by $p$. At the receiving end, if $E + L > a$, then federate p will lag behind physical time by at least $E + L - a$. However, if $a > E + L$, it does not cause $p$'s logical time to lag behind physical time. The RTI, having authorized s to move to time $t$, cannot permit $p$ to advance to a time $t+a$ or beyond until it confirms the message's delivery to $p$. To mitigate risks associated with delays and ensure prompt processing of physical actions and meeting deadlines, it's advisable to set the after delay $a$ on connections to federates receiving network messages to exceed any anticipated $E+L$.

The centralized coordination approach ensures the precise and timely execution of activities and events across a federated network.

\subsection{Decentralized Coordination}
The decentralized coordination model extends PTIDES ~\cite{derler_ptides}. This model draws inspiration from works by Lamport, Chandy, and Misra \cite{chandy1979distributed, chandy1979distributed2}. In this decentralized coordination strategy, the RTI plays a limited role, coordinating startup, shutdown, and clock synchronization. It is not involved in the execution of the distributed program.

In this approach, each federate is associated with a Safe-to-Process (STP) offset defined by the user. For a given federate $f_i$, we define $S_i \in T$ as its STP offset. A federate is restricted from progressing to any event tag \(g=(t,m)\) until the condition \(Ti \geq t+S_i\) is satisfied, where \(T_i\) denotes the physical time on \(f_i\)'s machine. If \(f_i\) is associated with physical actions, then \(S_i \geq 0\). In other cases, $S_i$ may assume positive, negative, or zero values. The STP offset's purpose is to ensure that all potentially influencing events from other federates, with tags preceding g, are received by $f_i$ by the time the physical clock fulfills the aforementioned condition, thus facilitating processing in tag order.

Federates communicate directly through sockets in a peer-to-peer architecture, bypassing the RTI, and logical time advancement does not require RTI to be involved. Federates can proceed with their logical time to $t$ once their physical clock aligns with or after $t+STP$. Similar to the after clause, if the STP offset is greater than the total of network latency, clock synchronization error, and execution times combined, then every event will be handled in the order of their tags. Since the assumptions about network latency and others can be violated, HPRM also provides a handler for STP violation.

The decentralized coordination model is designed to prioritize availability over consistency. This makes it particularly suitable for applications like autonomous driving and other robotic systems interacting with dynamic environments that require responsive real-time behavior. These systems can often tolerate some inconsistencies, which can be carefully managed by the user through the STP offset in HPRM. The STP offset allows developers to fine-tune the balance between availability and consistency. Conversely, the centralized model ensures software components behave as specified, prioritizing consistency over availability. This model is more appropriate for safety-critical applications (e.g. aerospace systems and medical robots) that cannot tolerate any inconsistencies and where strict global ordering of events is required. Users can easily switch to decentralized coordination by specifying the target property in HPRM, which allows for flexible adaptation to varying applications.

\begin{figure}
    \centering
    \includegraphics[width=0.6\linewidth]{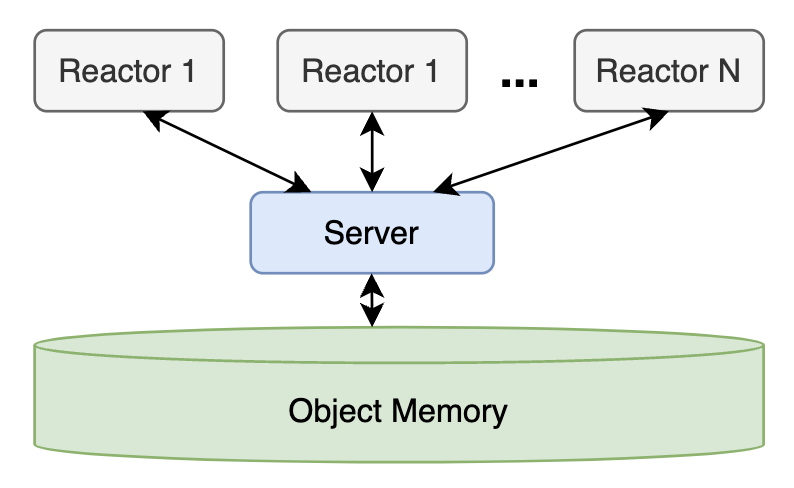}
    \caption{Architecture of in-memory object stores}
    \label{fig:objectstore}
    \vspace{-10pt}
\end{figure}

% \begin{figure*}[ht!]
%     \centering
%     % Adjust the width of minipages if necessary to fit your document layout.
%     \begin{minipage}{0.40\textwidth}
%         \centering
%         \includegraphics[width=\linewidth]{objectstore_small.png} % Ensure image is scaled to fit the minipage
%         \caption{Architecture of in-memory object stores}
%         \label{fig:objectstore}
%     \end{minipage}\hfill % This will fill up the space between the minipages
%     \begin{minipage}{0.48\textwidth}
%         \centering
%         \includegraphics[width=\linewidth]{plasma.png} % Ensure image is scaled to fit the minipage
%         \caption{Comparison of delay in writing NumPy arrays to shared memory between Python shared memory module and Plasma object store}
%         \label{fig:plasma}
%     \end{minipage}
% \end{figure*}

Previous research ~\cite{10316195} has shown that even under minimal stress on ROS2, it can observe dangerous out-of-order message sequences 0.2\% of the time (600 out of 300k tests). This error rate increases by two orders of magnitude under stress. In contrast, it has been verified that using centralized coordination implementation yields zero errors over 300k test runs for this scenario. Using decentralized coordination in LF, no errors are found for realistic message publishing periods down to 1ms. Errors only began appearing for unrealistically small periods below 1ms, but unlike ROS2, these errors were detectable.

\subsection{Optimizations}

\subsubsection{In-memory Object Store}
The shared memory (SM) module was introduced in Python 3.8 and has been used as a workaround to enable zero-copy in ROS2. By mapping the relevant region of shared memory into each process's address space, the module allows processes to access the same data without needing to copy data into separate buffers, thus saving CPU cycles and memory bandwidth. The Python SM module is used to create a block of shared memory that can be accessed by multiple processes. Processes can share complex data types more easily by using this shared memory block. It allows for the creation, destruction, and management of shared memory segments, and it supports the creation of NumPy arrays \cite{oliphant2006guide} that can directly map to a shared memory block. However, this approach is inefficient compared to the in-memory object stores when transferring large objects. 

In-memory object stores also enable zero-copy data transfer, reducing memory usage and improving performance. HPRM seamlessly integrates with the Plasma in-memory object store ~\cite{arrow2022cross}, automatically enabling it for the transfer of large payloads (greater than 64KB) between processes. This use of in-memory object store is inspired by Ray~\cite{moritz2018ray}. The architecture of Plasma object store is shown in Fig \ref{fig:objectstore}.  Plasma runs as a separate process and is written in C++ and is designed as a single-threaded event loop based on the Redis event loop library. The plasma client library can be linked into applications. Clients communicate with the Plasma store via messages serialized using Google Flatbuffers. 

Plasma outperforms Python's SM module for several reasons. Firstly, Plasma implements bulk eviction policies to manage memory more efficiently at scale. By evicting objects in bulk, the store can reduce the overhead associated with eviction, such as the cost of deciding which objects to evict and the process of eviction. Secondly, Plasma is designed to store data in a columnar format, which is optimized for efficient memory access and CPU cache utilization. This columnar format enables faster read and write operations, as well as better compression, leading to improved performance and reduced memory footprint compared to Python's SM module.

\begin{figure}
    \vspace{-15pt}
    \centering
    \includegraphics[width=0.6\linewidth]{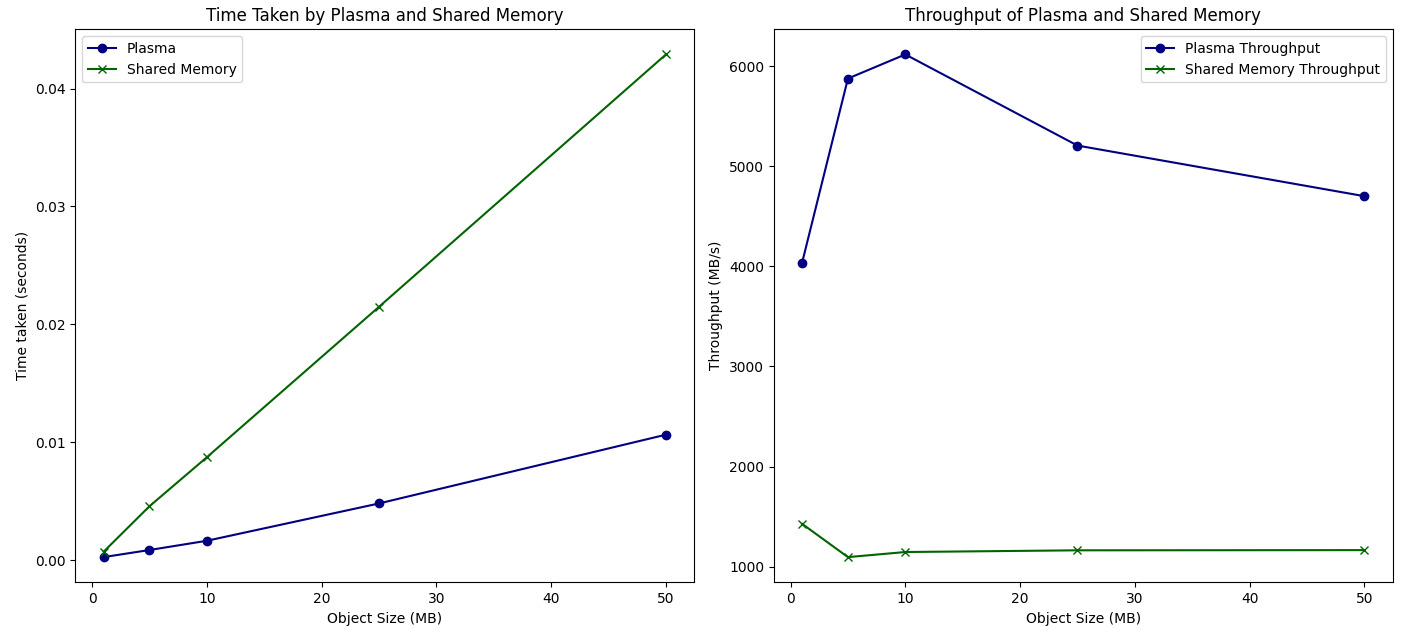}
    \caption{Comparison of delay in writing NumPy arrays to shared memory}
    \label{fig:plasma}
    \vspace{-10pt}
\end{figure}

Figure \ref{fig:plasma} compares the delay in milliseconds when writing NumPy arrays of varying sizes to shared memory, utilizing both the Python SM module and the Plasma object store. The Python SM module's delay appears to increase linearly with the object size, growing significantly faster than the Plasma object store's delay. In contrast, the Plasma object store's delay increases at a much slower rate as the size of the object grows. At the largest object size of 50 MB, the Python SM module's delay exceeds 40 milliseconds, while the Plasma object store's delay remains just under 10 milliseconds. This demonstrates the Plasma object store's superior performance and scalability when dealing with large objects, making it a more efficient choice for applications that require high-throughput data transfer between processes.

\subsubsection{Adaptive Serialization}
Traditional pickle serialization in Python often requires making one or more copies of the data being serialized. For example, when a large object is serialized, pickle first creates a bytes representation of the object, which is then written to the output stream. This process inherently involves copying the data. Out-of-band serialization, on the other hand, allows large data buffers to be handled separately from the main serialization stream. By using PickleBuffer objects, it's possible to avoid these additional memory copies, as the data does not need to be copied into the pickle stream but can instead be transmitted directly to the consumer in its original form.

By separating the metadata from the actual data buffers, out-of-band serialization allows the transmission of large data buffers without embedding them into the serialized pickle stream. This separation is particularly beneficial for applications that transmit large amounts of data (e.g. buffer-like objects, such as NumPy) between processes or over the network, as it enables the direct transfer of memory buffers without the overhead of serialization and deserialization processes.

\begin{figure}
    \centering
    \includegraphics[width=0.65\linewidth]{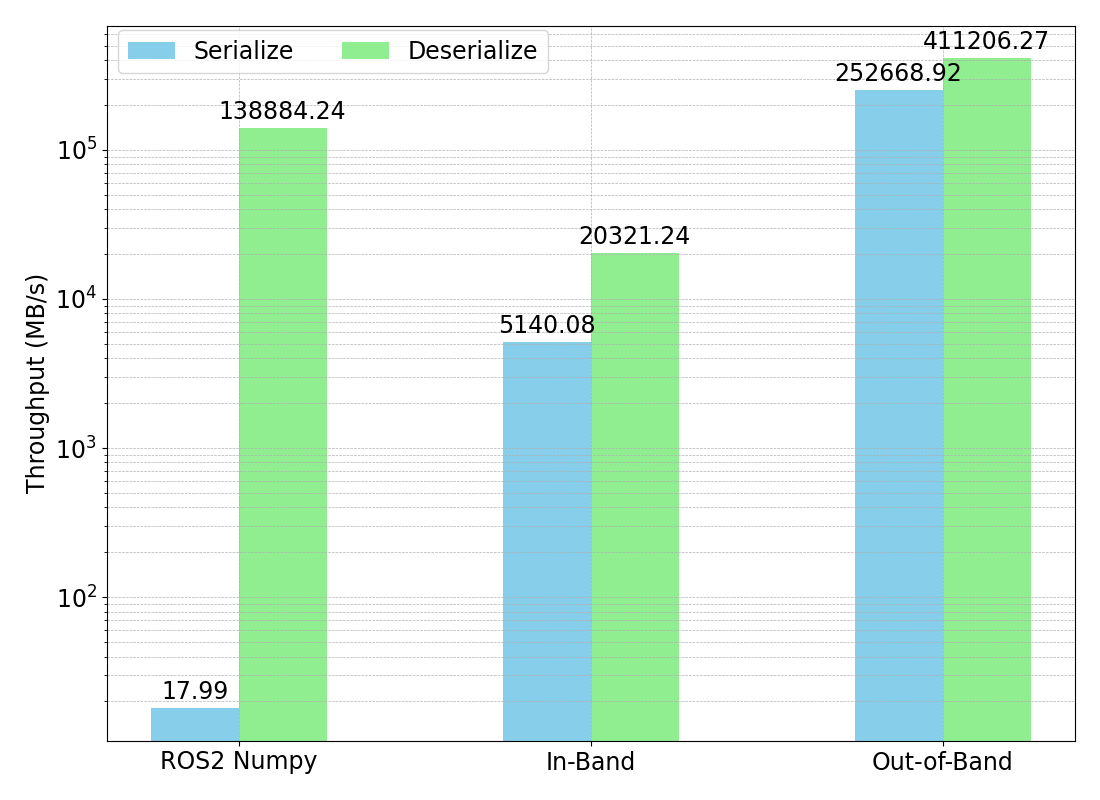}
     \caption{The serialization and deserialization throughput of ROS2 Numpy Package, in-band serialization, and out-of-band serialization}
    \label{fig:serialization}
    \vspace{-10pt}
\end{figure}

The benchmark shown in Figure ~\ref{fig:serialization} illustrates the serialization and deserialization throughput for a 5 MB NumPy Array. It compares the performance between ROS2 NumPy package\cite{ros2_numpy}, in-band serialization and out-of-band serialization. The ROS2 NumPy package serializes NumPy arrays by embedding them into a message data type supported by ROS2, such as PointCloud or Image. The results demonstrate that, compared to the ROS2 NumPy package and in-band serialization, out-of-band serialization achieves a serialization throughput that is 14,045x and 49.2x higher, and a deserialization throughput that is 2.96x and 20.2x greater, respectively.

HPRM implements adaptive serialization, which dynamically adjusts the serialization method based on the data type. Data types such as lists, byte arrays, and integers continue to be serialized using the in-band approach, while data types like NumPy arrays and Tensors utilize out-of-band serialization for optimal performance. Additionally, we have developed a recursive serializer for data structures storing different types of objects, including dictionary and list. This serializer separates objects like NumPy arrays from other objects that can be transmitted between reactors using regular serialization. The serialized bytes are then retrieved from the in-memory object store by the receiver and combined with the object received over the network to reconstruct the data structure.  

\subsubsection{Eager Protocol \& Real-Time Sockets}
\label{sec:realtimesocket}

To minimize the latency for transmitting small payloads, such as object references, metadata, and vehicle controls, we've implemented an eager protocol \cite{brightwell2003evaluation}. It pre-allocates fixed-size buffering space (64KB) for the message, reducing the handshake latency or wait time involved for the other federate to allocate memory for a new message. 

Also, the Nagle's algorithm \cite{minshall2000application}, enabled by default in TCP, bundles short messages together to avoid network traffic. As a result, it was delaying small messages. In the context of HPRM, disabling the Nagle's algorithm can be particularly beneficial when transmitting object references for large payloads stored in the in-memory object store. Without the Nagle's algorithm, these small but crucial references can be sent immediately, allowing the receiving process to access large payloads with minimal delay. 

\begin{figure}
    \centering
    \includegraphics[width=0.8\linewidth]{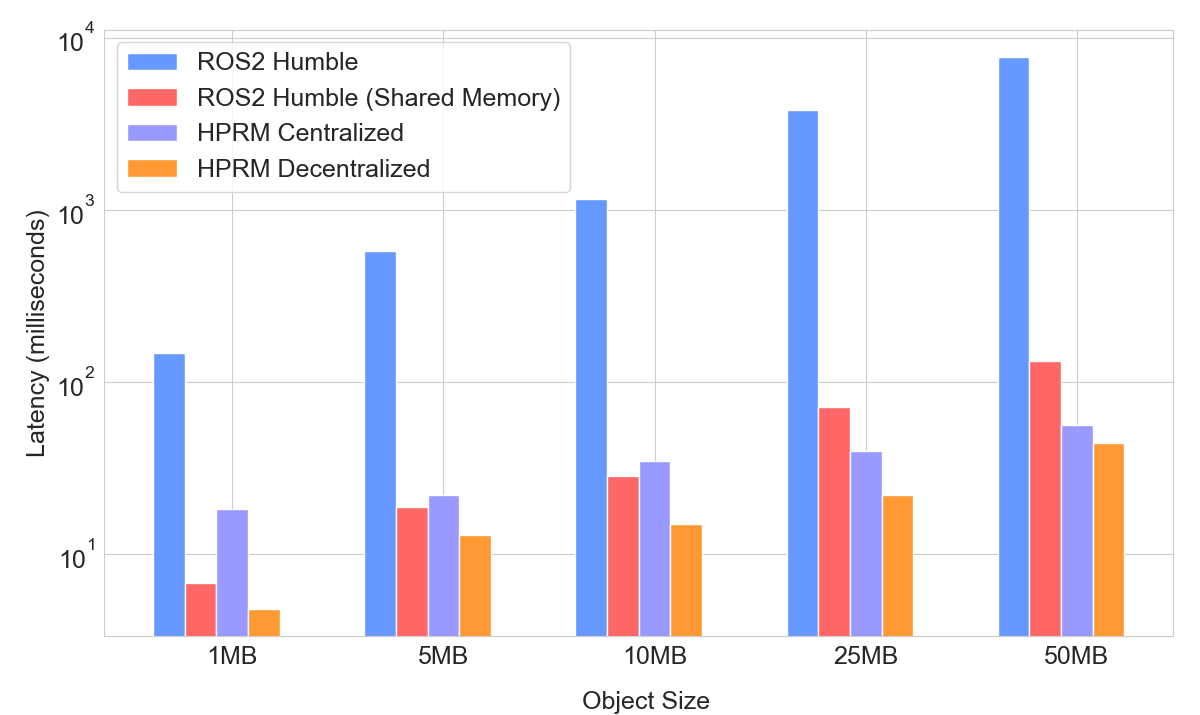}
    \caption{Mean latency of broadcast and gather on 4 nodes with different object sizes}
    \label{fig:latency}
    \vspace{-10pt}
\end{figure}

\section{Evaluation}
It is noted that all benchmarks were conducted on a personal workstation, equipped with an Intel$\textsuperscript{\textregistered}$ i7-13620H with 10 cores and a NVIDIA RTX4060 with 12GB GPU memory. We provide docker images for setting up the evaluation.
%The system runs on Ubuntu 22.04 and uses Python 3.10, ROS2 Humble, CARLA 0.9.15, PyTorch 2.2.2.

\subsection{Mean Latency}
Figure \ref{fig:latency} illustrates the comparison of average latency for broadcast and gather operations on objects with varying sizes using 4 nodes. This experiment utilizes ROS2 Humble, its variant with shared memory, and both the centralized and decentralized coordination strategy of HPRM. The x-axis represents the object size in megabytes (MB), ranging from 1 to 50 MB, while the y-axis indicates the latency in milliseconds, displayed on a logarithmic scale. The term latency refers to the duration required for one node to send a payload to another and for it to be received. This process may include sending and processing coordination-related messages, serialization and deserialization, and transferring data over the network. 

It is important to note that ROS2 Humble (Shared Memory) refers to using Python's pickle for serialization and the shared memory module IPC. This approach allows ROS2 to pass object references between processes, a workaround commonly utilized by robotics developers to leverage the benefits of shared memory. However, we argue that such a method is inefficient and requires users to consider the message size and type, as well as to manage the object read and write access manually. HPRM abstracts away these burdens for the users.

From the graph, we can observe that as the object size increases, the latency for both ROS2 and HPRM also increases. However, ROS2's latency grows at a higher rate than HPRM's. Specifically, for 10MB objects, the typical size of large camera images, ROS2's mean latency hits 1,161 ms, while HPRM's mean latency is around 15 ms—about 77x faster. For the largest object size of 50 MB, ROS2's average latency is at 7,723 ms, whereas HPRM's average latency is 44.6 ms, which is 173x faster. We also observed that the latency of decentralized coordination is lower than that of centralized coordination, as it prioritizes availability and incurs less synchronization overhead. Nevertheless, as the size of the object grows, the impact of synchronization overhead on the mean latency diminishes. We conclude that HPRM with decentralized coordination consistently shows the lowest latency across all object sizes in the plot, outperforming ROS2 Humble and its variant with shared memory.

\subsection{Applications}
To show the improvements in performance, we validated HPRM and ROS2 Humble on running reinforcement learning agents in the CARLA autonomous driving simulator. We designed a benchmark that simulates end-to-end urban driving, running ML models in parallel during inference. Specifically, we adapted a pre-trained Proximal Policy Optimization (PPO) agent developed by Zhang et al. \cite{zhang2021roach} to run in parallel with You Only Look Once (YOLO) \cite{Jocher_YOLOv5_by_Ultralytics_2020} for object detection.

% The neural network architecture used by the RL expert employs six convolutional layers to encode the bird’s-eye view (BEV) and two fully-connected (FC) layers to encode the measurement vector.  Outputs of both encoders are concatenated and then processed by another two FC layers to produce a latent feature, which is then fed into a value head and a policy head, each with two FC hidden layers. In conjunction, we utilize YOLOv5 for object detection. The YOLOv5 architecture is structured into three main components: the backbone, neck, and head. The backbone, leverages Cross-Stage Partial networks, extracts features from input images. The neck, built upon a Path Aggregation Network, processes these features to produce enriched feature maps at various scales. Finally, the head component uses these feature maps to predict bounding boxes and class probabilities for detected objects.

\begin{figure}
    \centering
    \includegraphics[width=\linewidth]{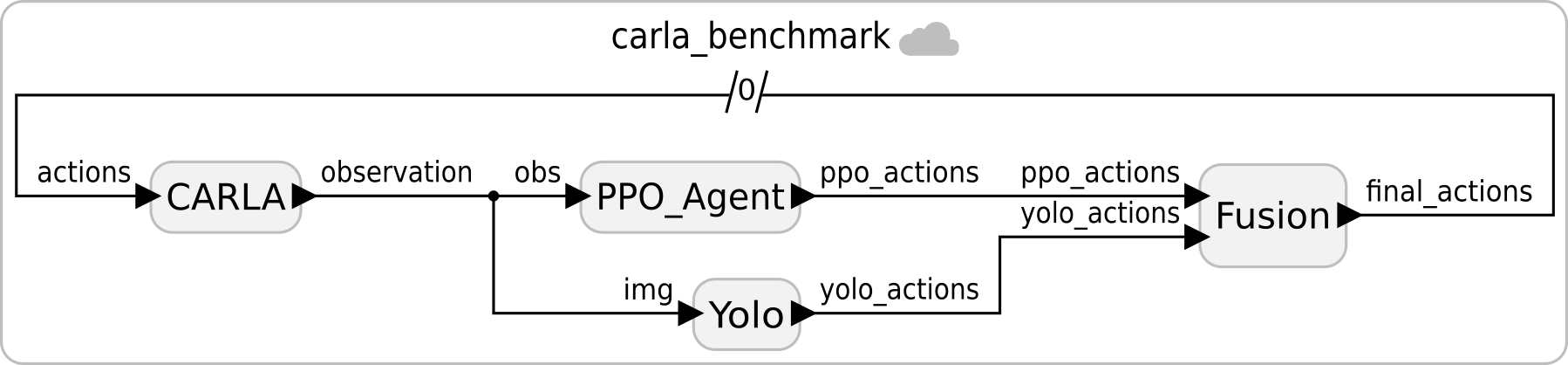}
    \caption{Dataflow diagram of the autonomous driving application}
    \label{fig:lf-diagram}
    \vspace{-10pt}
\end{figure}

Figure \ref{fig:lf-diagram} is a data flow diagram automatically generated by HPRM. The rendered BEV and RGB camera images from the CARLA simulator are passed separately to the PPO Agent reactor and the YOLO reactor. PPO Agent runs the policy and passes the policy action to Fusion reactor, while YOLO is executed in parallel and passes action based on object detection (e.g. STOP signs and traffic lights) to the Fusion reactor. The Fusion reactor processes the two actions and determines the final actions. When actions are received by the CARLA reactor, the simulator applies those actions, advancing to the next frame. The publisher and subscriber implementation of the benchmark in ROS2 follows the same paradigm as in HPRM, replacing reactors with ROS nodes. For synchronization purposes, the Fusion node in ROS2 blocks until it has received updated actions from both the PPO Agent and YOLO, after which it sends the final action to the CARLA node.
% It's worth noting that we used CARLA Python API instead of  ROS Bridge <TODO CITE> in our benchmarks due to version compatibility constraints.

\begin{figure}
    \centering
    \includegraphics[width=0.8\linewidth]{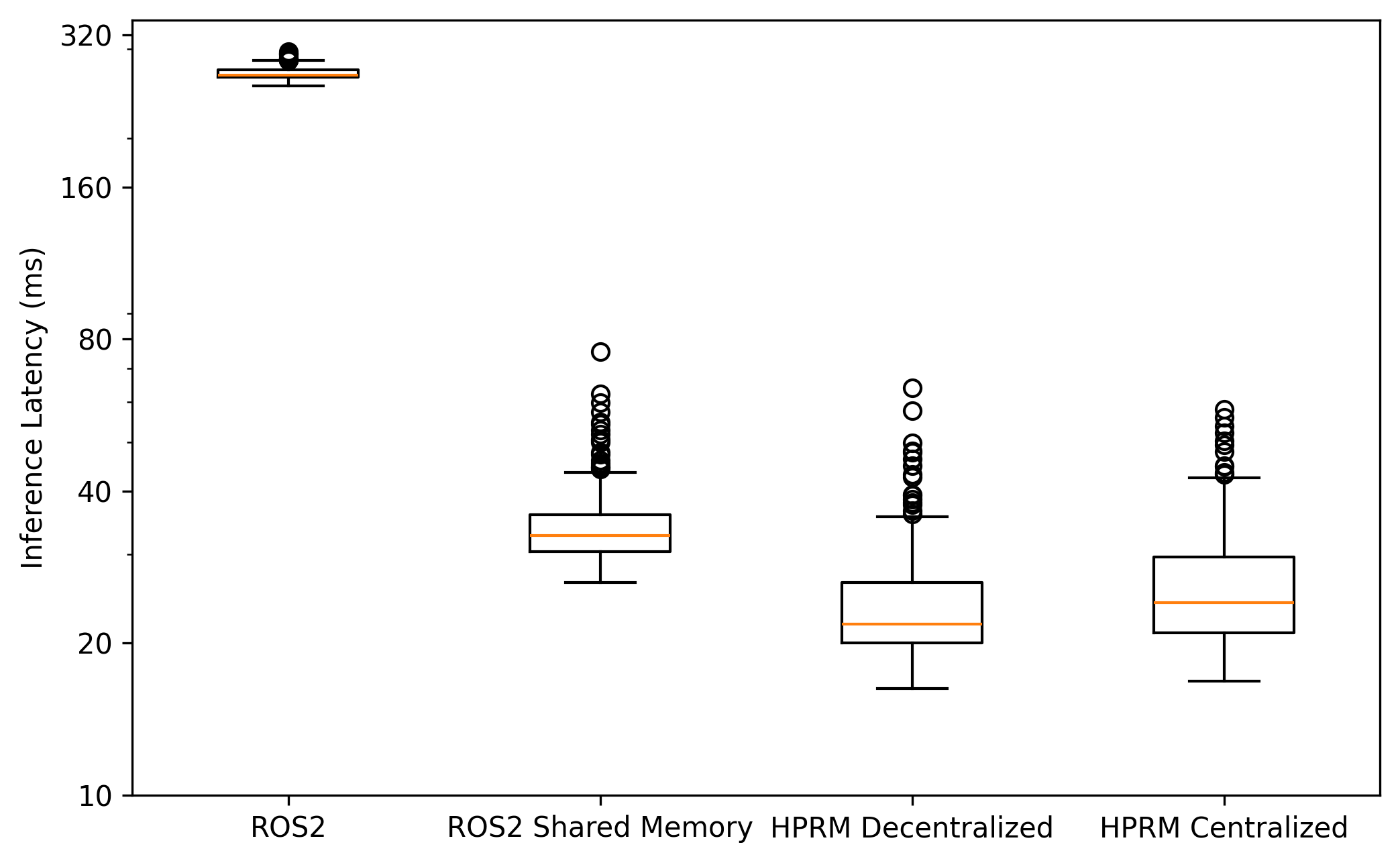}
    \caption{Performance of HPRM and ROS2 on the CARLA benchmark}
    \label{carla-performance}
    \vspace{-12pt}
\end{figure}

Inference latency is measured as the sum of communication time and inference time. We found that running PPO policy inference in CPU and YOLO in GPU led to a slight performance increase due to full utilization of compute resources, and was implemented across the benchmark. The box plot in Figure \ref{carla-performance} illustrates the inference latency measured when running the CARLA benchmark across 400 environment step frames after 100 warm-up steps with HPRM and ROS2 Humble. To obtain a more accurate measure of inference latency, our benchmarks exclude the time CARLA spends computing physics. When comparing HPRM decentralized coordination with the default ROS2 Humble implementation, the inference latency is reduced by 91.1\%. The box plot also shows that HPRM with decentralized and centralized coordination significantly outperforms ROS2 that uses shared memory, achieving latency reductions of 29.2\% and 23.6\% respectively. 

HPRM with decentralized coordination has the best performance with the note that frame rates, slightly outperforming HPRM with centralized coordination. The optimizations implemented in this research significantly lowered the IPC overhead, as further demonstrated in real-world application.

It's worth noting that the performance gap between HPRM and ROS2 is more pronounced in the latency benchmark than in the real-world application. This can be attributed to the fact that our optimizations primarily focus on reducing I/O overhead. The mean latency benchmark consists of more I/O-bound tasks compared to the actual application, where a significant portion of the computation time is spent on running inference for ML models. As a result, the impact of our optimizations is more evident in the mean latency benchmark, whereas the performance difference in the real-world application is relatively smaller.

\section{Conclusions}

In this paper, we presented HPRM, a high-performance robotic middleware designed to address the challenges of nondeterminism and high communication latency in intelligent autonomous systems. HPRM leverages a centralized and decentralized coordination model to ensure predictable event processing across distributed nodes. We introduced several optimizations in HPRM, including an in-memory object store for efficient zero-copy transfer of large payloads, adaptive serialization to minimize serialization overhead based on data types, and an eager protocol with real-time sockets to reduce handshake latency. Benchmark results showed that HPRM achieved up to 173x lower latency than ROS2 when transmitting large messages to multiple nodes. Furthermore, we validated the real-world applicability of HPRM by integrating it with the CARLA autonomous driving simulator and running deep reinforcement learning agents alongside object detection workloads. In this application, HPRM attained a 91.1\% lower latency than ROS2. Future work could investigate the scalability of HPRM in larger distributed robotic systems and explore its potential role in enabling efficient and deterministic communication for cloud robotics applications. In conclusion, HPRM represents a significant step forward in the development of deterministic, high-performance robotic middleware. 

\section{Acknowledgment}
The authors would like to thank Erling Rennemo Jellum for the contribution to the implementation of real-time sockets and anonymous reviewers for their helpful suggestions.

\addtolength{\textheight}{-2cm}   % This command serves to balance the column lengths

\bibliographystyle{IEEEtran}
\bibliography{references}

\begin{thebibliography}{10}
\providecommand{\url}[1]{#1}
\csname url@rmstyle\endcsname
\providecommand{\newblock}{\relax}
\providecommand{\bibinfo}[2]{#2}
\providecommand\BIBentrySTDinterwordspacing{\spaceskip=0pt\relax}
\providecommand\BIBentryALTinterwordstretchfactor{4}
\providecommand\BIBentryALTinterwordspacing{\spaceskip=\fontdimen2\font plus
\BIBentryALTinterwordstretchfactor\fontdimen3\font minus \fontdimen4\font\relax}
\providecommand\BIBforeignlanguage[2]{{%
\expandafter\ifx\csname l@#1\endcsname\relax
\typeout{** WARNING: IEEEtran.bst: No hyphenation pattern has been}%
\typeout{** loaded for the language `#1'. Using the pattern for}%
\typeout{** the default language instead.}%
\else
\language=\csname l@#1\endcsname
\fi
#2}}

\bibitem{sifakis2019autonomous}
J.~Sifakis, ``Autonomous systems--an architectural characterization,'' \emph{Models, Languages, and Tools for Concurrent and Distributed Programming: Essays Dedicated to Rocco De Nicola on the Occasion of His 65th Birthday}, pp. 388--410, 2019.

\bibitem{dinari2020inter}
H.~Dinari, ``Inter-process communication ({IPC}) in distributed environments: An investigation and performance analysis of some middleware technologies.'' \emph{International Journal of Modern Education \& Computer Science}, vol.~12, no.~2, 2020.

\bibitem{venkataraman2015evaluation}
A.~Venkataraman and K.~K. Jagadeesha, ``Evaluation of inter-process communication mechanisms,'' \emph{Architecture}, vol.~86, p.~64, 2015.

\bibitem{quigley2009ros}
M.~Quigley, K.~Conley, B.~Gerkey, J.~Faust, T.~Foote, J.~Leibs, R.~Wheeler, A.~Y. Ng, \emph{et~al.}, ``{ROS}: an open-source robot operating system,'' in \emph{ICRA workshop on open source software}, vol.~3, no. 3.2.\hskip 1em plus 0.5em minus 0.4em\relax Kobe, Japan, 2009, p.~5.

\bibitem{soni2017survey}
D.~Soni and A.~Makwana, ``A survey on {MQTT}: a protocol of internet of things ({IoT}),'' in \emph{International conference on telecommunication, power analysis and computing techniques (ICTPACT-2017)}, vol.~20, 2017, pp. 173--177.

\bibitem{hunt2013ddos}
N.~Hunt, T.~Bergan, L.~Ceze, and S.~D. Gribble, ``{DDOS}: taming nondeterminism in distributed systems,'' \emph{ACM SIGPLAN Notices}, vol.~48, no.~4, pp. 499--508, 2013.

\bibitem{blass2021automatic}
T.~Blass, A.~Hamann, R.~Lange, D.~Ziegenbein, and B.~B. Brandenburg, ``Automatic latency management for {ROS2}: Benefits, challenges, and open problems,'' in \emph{2021 IEEE 27th Real-Time and Embedded Technology and Applications Symposium (RTAS)}.\hskip 1em plus 0.5em minus 0.4em\relax IEEE, 2021, pp. 264--277.

\bibitem{lohstroh2021toward}
M.~Lohstroh, C.~Menard, S.~Bateni, and E.~A. Lee, ``Toward a lingua franca for deterministic concurrent systems,'' \emph{ACM Transactions on Embedded Computing Systems (TECS)}, vol.~20, no.~4, pp. 1--27, 2021.

\bibitem{lohstroh2020reactors}
M.~Lohstroh, {\'I}.~{\'I}. Romeo, A.~Goens, P.~Derler, J.~Castrillon, E.~A. Lee, and A.~Sangiovanni-Vincentelli, ``Reactors: A deterministic model for composable reactive systems,'' in \emph{Cyber Physical Systems. Model-Based Design: 9th International Workshop, CyPhy 2019, and 15th International Workshop, WESE 2019, New York City, NY, USA, October 17-18, 2019, Revised Selected Papers 9}.\hskip 1em plus 0.5em minus 0.4em\relax Springer, 2020, pp. 59--85.

\bibitem{puck2021performance}
L.~Puck, P.~Keller, T.~Schnell, C.~Plasberg, A.~Tanev, G.~Heppner, A.~Roennau, and R.~Dillmann, ``Performance evaluation of real-time {ROS2} robotic control in a time-synchronized distributed network,'' in \emph{2021 IEEE 17th International Conference on Automation Science and Engineering (CASE)}.\hskip 1em plus 0.5em minus 0.4em\relax IEEE, 2021, pp. 1670--1676.

\bibitem{dosovitskiy2017carla}
A.~Dosovitskiy, G.~Ros, F.~Codevilla, A.~Lopez, and V.~Koltun, ``Carla: An open urban driving simulator,'' in \emph{Conference on robot learning}.\hskip 1em plus 0.5em minus 0.4em\relax PMLR, 2017, pp. 1--16.

\bibitem{fitzpatrick2014middle}
P.~Fitzpatrick, E.~Ceseracciu, D.~E. Domenichelli, A.~Paikan, G.~Metta, and L.~Natale, ``A middle way for robotics middleware,'' \emph{Journal of Software Engineering for Robotics}, vol.~5, no.~2, pp. 42--49, 2014.

\bibitem{ros2_rosidl_issue_156}
``Extremely slow message creation for large arrays in python,'' \url{https://github.com/ros2/rosidl_python/issues/156}.

\bibitem{ros2_rclpy_issue_763}
``Publishing large data is 30x-100x slower than for rclcpp,'' \url{https://github.com/ros2/rclpy/issues/763}.

\bibitem{ros2_issue_1242}
``Very slow publishing of large messages,'' \url{https://github.com/ros2/ros2/issues/1242}.

\bibitem{kronauer2021latency}
T.~Kronauer, J.~Pohlmann, M.~Matth{\'e}, T.~Smejkal, and G.~Fettweis, ``Latency analysis of {ROS2} multi-node systems,'' in \emph{2021 IEEE International Conference on Multisensor Fusion and Integration for Intelligent Systems (MFI)}.\hskip 1em plus 0.5em minus 0.4em\relax IEEE, 2021, pp. 1--7.

\bibitem{pardo2003omg}
G.~Pardo-Castellote, ``Omg data-distribution service: Architectural overview,'' in \emph{23rd International Conference on Distributed Computing Systems Workshops, 2003. Proceedings.}\hskip 1em plus 0.5em minus 0.4em\relax IEEE, 2003, pp. 200--206.

\bibitem{ros2_rclpy_issue_833}
``Memory leak in subscription when using zero-copy with rmw\_cyclonedds,'' \url{https://github.com/ros2/rclpy/issues/833}.

\bibitem{liu2020robotic}
W.~Liu, H.~Wu, Z.~Jiang, Y.~Gong, and J.~Jin, ``A robotic communication middleware combining high performance and high reliability,'' in \emph{2020 IEEE 32nd International Symposium on Computer Architecture and High Performance Computing (SBAC-PAD)}.\hskip 1em plus 0.5em minus 0.4em\relax IEEE, 2020, pp. 217--224.

\bibitem{liu2022zoro}
W.~Liu, J.~Jin, H.~Wu, Y.~Gong, Z.~Jiang, and J.~Zhai, ``Zoro: A robotic middleware combining high performance and high reliability,'' \emph{Journal of Parallel and Distributed Computing}, vol. 166, pp. 126--138, 2022.

\bibitem{wang2019tzc}
Y.-P. Wang, W.~Tan, X.-Q. Hu, D.~Manocha, and S.-M. Hu, ``Tzc: Efficient inter-process communication for robotics middleware with partial serialization,'' in \emph{2019 IEEE/RSJ International Conference on Intelligent Robots and Systems (IROS)}.\hskip 1em plus 0.5em minus 0.4em\relax IEEE, 2019, pp. 7805--7812.

\bibitem{lee2023consistency}
E.~A. Lee, R.~Akella, S.~Bateni, S.~Lin, M.~Lohstroh, and C.~Menard, ``Consistency vs. availability in distributed cyber-physical systems,'' \emph{ACM Transactions on Embedded Computing Systems}, vol.~22, no.~5s, pp. 1--24, 2023.

\bibitem{menard2023high}
C.~Menard, M.~Lohstroh, S.~Bateni, M.~Chorlian, A.~Deng, P.~Donovan, C.~Fournier, S.~Lin, F.~Suchert, T.~Tanneberger, \emph{et~al.}, ``High-performance deterministic concurrency using lingua franca,'' \emph{ACM Transactions on Architecture and Code Optimization}, vol.~20, no.~4, pp. 1--29, 2023.

\bibitem{kwok2024efficient}
\BIBentryALTinterwordspacing
J.~Kwok, M.~Lohstroh, and E.~A. Lee, ``Efficient parallel reinforcement learning framework using the reactor model,'' in \emph{Proceedings of the 36th ACM Symposium on Parallelism in Algorithms and Architectures}, ser. SPAA '24.\hskip 1em plus 0.5em minus 0.4em\relax New York, NY, USA: Association for Computing Machinery, 2024, p. 41–51. [Online]. Available: \url{https://doi.org/10.1145/3626183.3659967}
\BIBentrySTDinterwordspacing

\bibitem{derler_ptides}
\BIBentryALTinterwordspacing
P.~Derler, T.~H. Feng, E.~A. Lee, S.~Matic, H.~D. Patel, Y.~Zhao, and J.~Zou, ``Ptides: A programming model for distributed real-time embedded systems,'' May 2008. [Online]. Available: \url{http://www2.eecs.berkeley.edu/Pubs/TechRpts/2008/EECS-2008-72.html}
\BIBentrySTDinterwordspacing

\bibitem{chandy1979distributed}
K.~M. Chandy and J.~Misra, ``Distributed simulation: A case study in design and verification of distributed programs,'' \emph{IEEE Transactions on software engineering}, no.~5, pp. 440--452, 1979.

\bibitem{chandy1979distributed2}
K.~M. Chandy, V.~Holmes, and J.~Misra, ``Distributed simulation of networks,'' \emph{Computer Networks (1976)}, vol.~3, no.~2, pp. 105--113, 1979.

\bibitem{10316195}
S.~Bateni, M.~Lohstroh, H.~S. Wong, H.~Kim, S.~Lin, C.~Menard, and E.~A. Lee, ``Risk and mitigation of nondeterminism in distributed cyber-physical systems,'' in \emph{2023 21st ACM-IEEE International Symposium on Formal Methods and Models for System Design (MEMOCODE)}, 2023, pp. 1--11.

\bibitem{oliphant2006guide}
T.~E. Oliphant \emph{et~al.}, \emph{Guide to numpy}.\hskip 1em plus 0.5em minus 0.4em\relax Trelgol Publishing USA, 2006, vol.~1.

\bibitem{arrow2022cross}
A.~Arrow, ``A cross-language development platform for in-memory data,'' 2022.

\bibitem{moritz2018ray}
P.~Moritz, R.~Nishihara, S.~Wang, A.~Tumanov, R.~Liaw, E.~Liang, M.~Elibol, Z.~Yang, W.~Paul, M.~I. Jordan, \emph{et~al.}, ``Ray: A distributed framework for emerging $\{$AI$\}$ applications,'' in \emph{13th USENIX symposium on operating systems design and implementation (OSDI 18)}, 2018, pp. 561--577.

\bibitem{ros2_numpy}
``{ROS2} {NumPy}, {Tooling} for converting {ROS} messages to and from {NumPy} arrays,'' \url{https://github.com/Box-Robotics/ros2_numpy}.

\bibitem{brightwell2003evaluation}
R.~Brightwell and K.~Underwood, ``Evaluation of an eager protocol optimization for mpi,'' in \emph{European Parallel Virtual Machine/Message Passing Interface Users’ Group Meeting}.\hskip 1em plus 0.5em minus 0.4em\relax Springer, 2003, pp. 327--334.

\bibitem{minshall2000application}
G.~Minshall, Y.~Saito, J.~C. Mogul, and B.~Verghese, ``Application performance pitfalls and tcp's nagle algorithm,'' \emph{ACM SIGMETRICS Performance Evaluation Review}, vol.~27, no.~4, pp. 36--44, 2000.

\bibitem{zhang2021roach}
Z.~Zhang, A.~Liniger, D.~Dai, F.~Yu, and L.~Van~Gool, ``End-to-end urban driving by imitating a reinforcement learning coach,'' in \emph{Proceedings of the IEEE/CVF International Conference on Computer Vision (ICCV)}, 2021.

\bibitem{Jocher_YOLOv5_by_Ultralytics_2020}
\BIBentryALTinterwordspacing
G.~Jocher, ``{YOLOv5 by Ultralytics},'' May 2020. [Online]. Available: \url{https://github.com/ultralytics/yolov5}
\BIBentrySTDinterwordspacing

\end{thebibliography}
\end{document}